\documentclass[10pt, a4paper]{article}

\usepackage[]{lrec2026}
\usepackage{graphicx}
\usepackage{comment}
\usepackage{tabularx}
\usepackage{makecell}
\usepackage{multicol, multirow}
\usepackage{xcolor}
\usepackage{booktabs}
\usepackage{hyperref}
\usepackage{subfig}

\begin{document}

\title{ENEIDE: A High Quality Silver Standard Dataset for Named Entity Recognition and Linking in Historical Italian}

\name{Cristian Santini$^{a,c}$, Sebastian Barzaghi$^{b}$, Paolo Sernani$^{a}$, \\ 
{\bf \large Emanuele Frontoni$^{a}$, Laura Melosi$^{a}$, Mehwish Alam$^{c}$}\\}

\address{$^{a}$University of Macerata, Macerata, Italy \\
        $^{b}$University of Bologna, Bologna, Italy \\
        $^{c}$Telecom Paris, Polytechnic Institute of Paris, Palaiseau, France\\\\
         sebastian.barzaghi2@unibo.it, mehwish.alam@telecom-paris.fr\\
         \{c.santini12, paolo.sernani, emanuele.frontoni, laura.melosi\}@unimc.it\\}

\abstract{
This paper introduces ENEIDE (Extracting Named Entities from Italian Digital Editions), a silver standard dataset for Named Entity Recognition and Linking (NERL) in historical Italian texts. The corpus comprises 2,111 documents with over 8,000 entity annotations semi-automatically extracted from two scholarly digital editions: Digital Zibaldone, the philosophical diary of the Italian poet Giacomo Leopardi (1798--1837), and Aldo Moro Digitale, the complete works of the Italian politician Aldo Moro (1916--1978). Annotations cover multiple entity types (person, location, organization, literary work) linked to Wikidata identifiers, including NIL entities that cannot be mapped to the knowledge graph. To the best of our knowledge, ENEIDE represents the first multi-domain, publicly available NERL dataset for historical Italian with training, development, and test splits. We present a methodology for semi-automatic annotations extraction from manually curated scholarly digital editions, including quality control and annotation enhancement procedures. Baseline experiments using state-of-the-art models demonstrate the dataset's challenge for NERL and the gap between zero-shot approaches and fine-tuned models. The dataset's diachronic coverage spanning two centuries makes it particularly suitable for temporal entity disambiguation and cross-domain evaluation. ENEIDE is released under a CC BY-NC-SA 4.0 license.
 \\ \newline \Keywords{named entity recognition, entity linking, digital humanities, historical Italian, scholarly digital editions, Wikidata}\\ }

\maketitleabstract

\section{Introduction}

While the majority of Natural Language Processing (NLP) systems are trained and evaluated on contemporary born-digital texts, historical documents (including letters, printed books, and manuscripts) remain fundamental sources for scholars seeking to validate and enrich our understanding of historical events and cultural heritage. Despite their importance in the Digital Humanities field (DH), State-of-The-Art (SoTA) systems for Named Entity Recognition (NER) and Entity Linking (EL) (together NERL) demonstrate significant performance degradation~\citep{survey_hist_ner_2023} when applied to historical documents. This degradation stems from multiple factors: the linguistic complexity and variation of historical texts, the presence of noise introduced by Optical Character Recognition (OCR), and the scarce representation of historical language varieties in modern web-crawled corpora used to train contemporary language models.

The development of robust historical NERL systems, which enable the detection of textual references to entities (such as persons, locations, and organizations) and their disambiguation through Knowledge Graphs (KGs) like Wikidata~\citep{vrandevcic2014wikidata}, critically depends on the availability of annotated historical corpora for training and evaluation. However, the creation of such resources faces substantial obstacles. First, the field lacks widely adopted community standards for annotation schemas and guidelines, limiting interoperability and reusability across projects~\citep{survey_hist_ner_2023}. Second, the time and cost required for expert human annotation create a significant bottleneck, leaving many historical language varieties severely under-resourced. Historical Italian exemplifies this challenge: currently, only one dataset, i.e., MHERCL~\citep{graciotti-etal-2025-ke-mhisto}, provides NERL annotations for historical Italian texts. However, this resource is domain-specific (music periodicals from the 20th century), contains only a test split, and therefore offers limited utility for training purposes.

In this landscape, Scholarly Digital Editions (SDEs)~\citep{sahle_what_2016} emerge as a promising yet underutilized resource. SDEs are digital resources that provide manually curated annotations of named entities in TEI/XML format, created by domain experts. They play a pivotal role in broadening the scope of NERL research in historical texts and enhancing the scalability of NLP applications. These resources contain manual annotations made by domain-experts for people, places, organizations, bibliographic resources, and other entities referenced in texts from different historical periods, typically disambiguated using Wikidata or other domain-agnostic authority files. Reusing such datasets can improve and verify the application of historical NERL systems due to the validity of the interpretative work of philologists and expert scholars~\citep{valette2024does}. However, the systematic extraction and transformation of SDE annotations into reusable datasets remains an open methodological challenge.

This paper addresses this gap by presenting \textbf{ENEIDE} (\textit{Extracting Named Entities from Italian Digital Editions}), a silver standard dataset for NERL in historical Italian, semi-automatically extracted from two Italian SDEs: DigitalZibaldone\footnote{\href{https://digitalzibaldone.net/}{digitalzibaldone.net}}~\citep{stoyanova_remediating_2014}, based on the philosophical diary of the Italian poet Giacomo Leopardi (1798--1837), and Aldo Moro Digitale\footnote{\href{https://aldomorodigitale.unibo.it/}{aldomorodigitale.unibo.it}}~\citep{barzaghi_amd_2025}, containing the works of the Italian politician Aldo Moro (1916--1978). 

We refer to ENEIDE as a silver standard dataset to distinguish it from gold standard resources~\citep{wissler2014gold}, where all annotations are manually created and validated by expert annotators from scratch. Silver standard datasets are instead produced through semi-automatic methods~\citep{rebholz-schuhmann-etal-2010-calbc} — such as the extraction and transformation of pre-existing annotations, automated labeling, or distant supervision — followed by partial human validation. While silver standard resources may contain residual noise compared to fully manual gold annotations, they offer a scalable and cost-effective alternative for producing training data in low-resource scenarios, where gold standard annotation would be prohibitively expensive or time-consuming.

ENEIDE comprises 2,111 documents spanning two centuries and two distinct domains, with over 8,000 entity annotations across multiple types (person, location, organization, literary work) linked to Wikidata identifiers. To the best of our knowledge, ENEIDE represents the first multi-domain, publicly available dataset for NERL in historical Italian with training, development, and test splits. 

This work offers three key contributions:

\begin{enumerate}

\item A discussion of the methodology for semi-automatic dataset extraction from SDEs, detailing sampling strategies, quality control procedures, and annotation enhancement techniques;

\item The description of ENEIDE, a multi-domain, diachronic corpus covering Italian philosophical, literary, political, and legal texts from the 19th and 20th centuries semi-automatically extracted from SDEs;

\item An empirical analysis of dataset quality and characteristics, including average token counts, entity distribution statistics, temporal coverage, and NIL entity analysis.
\end{enumerate}

The remainder of this paper is structured as follows. Section~\ref{sec:related} surveys related work on historical NERL datasets. Section~\ref{sec:dataset} details our semi-automatic extraction pipeline and quality control procedures, while also offering dataset statistics. Section~\ref{sec:experiments} provides baseline experiments and analysis demonstrating the dataset's utility. Section~\ref{sec:discussion} discusses the dataset's reuse potential across different research scenarios. Section~\ref{sec:conclusion} summarizes our contributions and outlines future directions. Finally, Section~\ref{sec:limitations} addresses current limitations and potential solutions for dataset improvement.

The dataset is released under a CC BY-NC-SA 4.0 license on Zenodo~\citep{ENEIDE}.\footnote{\href{https://doi.org/10.5281/zenodo.17407356}{doi.org/10.5281/zenodo.17407356}}

\section{Related Work}
\label{sec:related}

The challenges of adapting NERL approaches to documents from cultural heritage institutions have led to the creation of several annotated historical corpora. This section discusses the most relevant datasets containing literary and political texts, with particular attention to resources for Italian. A comprehensive survey of historical NERL datasets is available in~\cite{survey_hist_ner_2023}.

\paragraph{Literary and Multilingual Corpora}
A seminal work in NER for literary texts was presented by~\citet{bamman_annotated_2019}. Their dataset, \textit{LitBank}, serves as a training and evaluation resource for multiple information extraction tasks: NER, coreference resolution, and event detection. The dataset comprises text samples from 100 English novels from Project Gutenberg. The annotations follow the ACE guidelines~\citep{walker_christopher_ace_2006}, including six coarse-grained entity categories (person, location, geo-political entity, facility, organization, and vehicle) and covering both common nouns and nested entities.

More recently,~\citet{ajmc_2024} presented a manually annotated dataset of named entity references in classical commentaries of Sophocles' \textit{Ajax}. The dataset, called \textit{AJMC}, was included in the HIPE-2022 shared task~\citep{ehrmann2022overview} and supports training and evaluation of systems for detecting general entities (persons, locations, and dates) and domain-specific entities (primary and secondary literary sources), along with their disambiguation using Wikidata. Entities are annotated according to 6 coarse-grained and 9 fine-grained classes, include nested entities at 1-level depth, and cover OCR-ed texts in three languages: English, French, and German. Similar to ENEIDE, entities are often referenced through abbreviations, such as "Cic." for "Cicero", a common practice in philological texts like classical commentaries and literary works such as Leopardi's Zibaldone.

Regarding literary entities specifically, one of the earliest datasets focusing on references to books, monographs, and essays is \textit{LinkedBooks}~\citep{colavizza_annotated_2017}. Extracted from documents related to Venetian historiography in multiple languages (including Italian), this dataset contains numerous annotated citations from reference lists and footnotes, including abbreviated primary and secondary sources. While this resource was designed to train NER models capable of extracting and parsing literary work references, it does not include general entity types (persons, locations, organizations) and lacks Wikidata identifiers.

\paragraph{Italian Historical Corpora}
For Italian specifically, a pivotal contribution was made by~\citet{paccosi-palmero-aprosio-2022-kind}, who describe \textit{KIND}, a multi-domain NER dataset extracted from various text types, including news, literary texts, and political works. Like ENEIDE, this dataset includes excerpts from the digital edition of Aldo Moro's works. The resource contains semi-automatic annotations using three coarse-grained entity types: person, location, and organization. However, the annotations do not consider nested entities, and the dataset does not support entity disambiguation through Wikidata.

Most recently,~\citet{graciotti-etal-2025-ke-mhisto} introduced MHERCL, the first dataset providing NERL annotations for historical Italian and English texts. This resource is extracted from a domain-specific corpus of 19th-century music periodicals. Despite its rich tagset and substantial annotations of long-tail (i.e., unpopular) entities, MHERCL contains only a test split, limiting its utility for training purposes.

\paragraph{Positioning ENEIDE}
ENEIDE differs from these existing resources in several key aspects: (1) it is the first multi-domain corpus for NERL in historical Italian with training, development, and test splits; (2) it combines general entity types (person, location, organization) with domain-specific ones (literary works); and (3) it covers two distinct historical periods (19th and 20th centuries) and textual genres (philosophical-literary and political-legal), providing temporal broadness uncommon in existing Italian resources.

\section{ENEIDE Dataset}
\label{sec:dataset}

The ENEIDE dataset spans multiple time periods and domains. It was semi-automatically extracted from two SDEs: Digital Zibaldone (DZ) and Aldo Moro Digitale (AMD). These sources cover two centuries (19th and 20th) and include various text types, making them suitable for testing EL systems on humanistic documents with complex contextual and historical features.

DZ is the digital edition of Giacomo Leopardi's \textit{Zibaldone di pensieri}, containing over 4,500 pages of reflections on literature, history, and philosophy written between 1817 and 1832. The hypertextual structure of this work, composed of cross-references (internal links between notes) as well as external ones (bibliographic references, quotes, named entities, etc.) motivated digital scholars to re-mediate this literary work into a digital research platform which would enable users to dynamically mine the text's structural complexity through XML/TEI~\cite{stoyanova_working_2023}.

The digital edition, collaboratively annotated by domain experts at Princeton University, serves as a valuable tool for assessing entity disambiguation methods for two main reasons. First, it includes thousands of expert-annotated references to people (\texttt{PER}), locations (\texttt{LOC}), and bibliographic works (\texttt{WORK}), linked when possible to Wikidata or VIAF. Second, Leopardi's notes offer a pre-digital example of an encyclopedic hypertext, comprising thousands of external references to historical figures, literary works, authors, and various facets of humanistic knowledge~\cite{stoyanova_fragmentary_2013}.

\begin{figure*}
    \centering
    \includegraphics[width=1\textwidth]{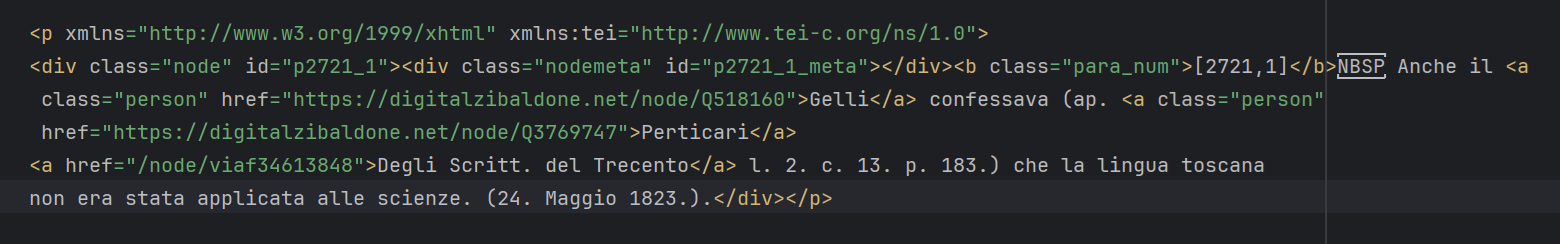}
    \caption{Example of markup with entities annotated for a note in Digital Zibaldone.}
    \label{fig:digitalzibaldone}
\end{figure*}

The structure of this resource resembles that of the original work, with every fragment available in a single web page as an HTML document, ordered by its page and identified by a unique URI. For instance, accessing the URI \url{https://digitalzibaldone.net/node/p2721_1} allows users to view the first note on page 2721 of the Zibaldone. References to people, places, and bibliographic works are formatted as hyperlinks directing users to a unique identifier. The unique identifiers are based mostly on Wikidata or VIAF if a corresponding Wikidata entry does not exist. The content of the platform is publicly available under a CC-BY-NC-SA. For this project we used as source the HTML documents available inside the digital platform. A sample of an HTML file contained in the digital edition is available in Figure~\ref{fig:digitalzibaldone}. 

\begin{figure*}
  \centering
  \includegraphics[width=1\textwidth]{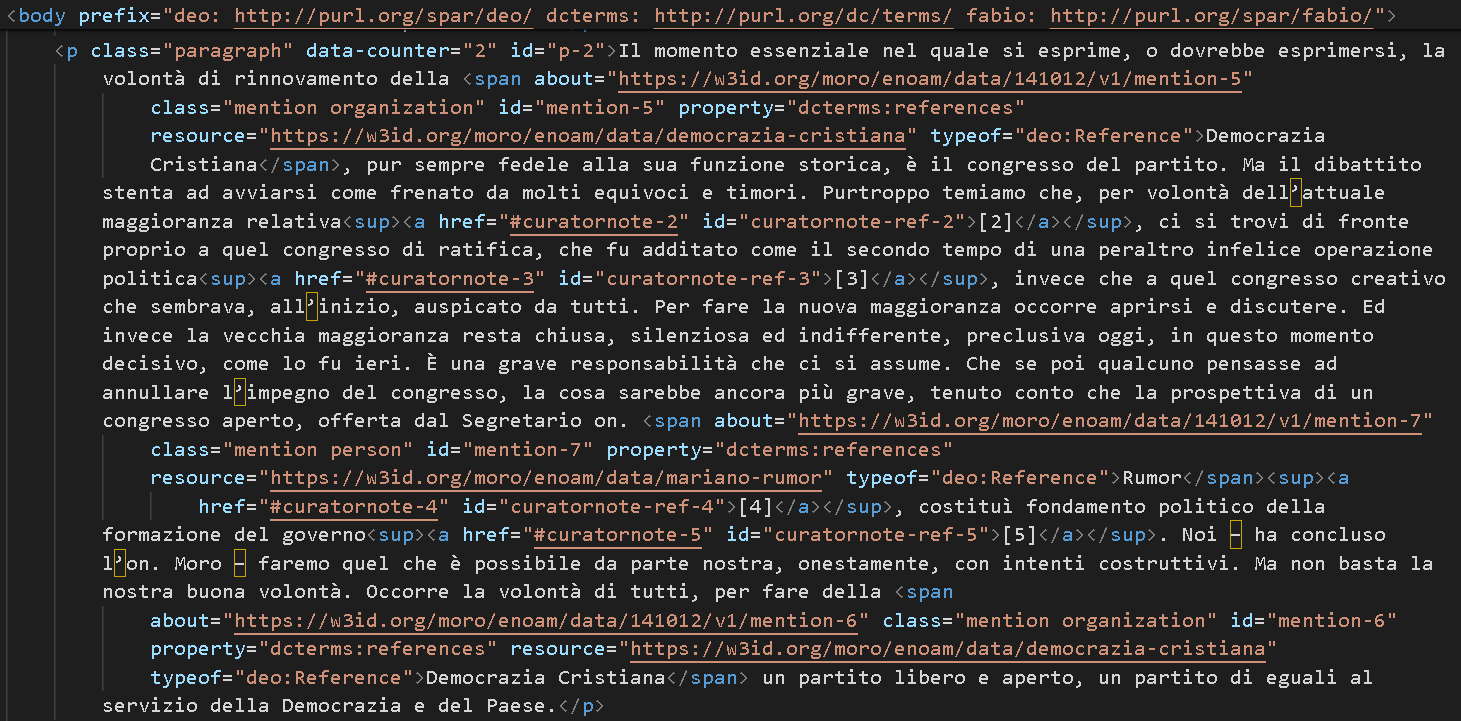}
  \caption{Example of entities annotated with HTML and RDFa in a document of the Edition of Aldo Moro's works.}
  \label{fig:sample_aldomoro}
\end{figure*}

AMD presents the complete works of Aldo Moro, collecting political, legal, and journalistic texts from the 1932 to 1978. Encoded in RDFa~\cite{adida2007rdfa} with semantic web annotations, AMD links three entity types to Wikidata: person (\texttt{PER}), location (\texttt{LOC}), and organization (\texttt{ORG}). Figure~\ref{fig:sample_aldomoro} illustrates an example of how RDFa is used to inject semantic information in HTML elements representing mentioned entities in the text.

Both corpora present notable challenges for NERL systems: in Aldo Moro, entities may be referenced indirectly (e.g., "Pope" for "Pope Pius XII"), while in Leopardi's texts complex abbreviations (e.g., "Il." for "Iliad") and metonymic references are abundant. Moreover, some entities do not exist in Wikidata. This last category requires NIL prediction: the ability to identify entities that cannot be mapped to any KG entry.

\subsection{Annotation Extraction and Sampling Strategy}

We extracted entities using \texttt{Beautiful Soup}\footnote{\href{https://pypi.org/project/beautifulsoup4/}{pypi.org/project/beautifulsoup4}}, a Python library for parsing HTML and XML documents, identifying annotated entities through HTML hyperlinks. For AMD, we first leveraged its public API\footnote{\href{https://aldomorodigitale.unibo.it/api/works}{aldomorodigitale.unibo.it/api/works}} to download all works in HTML format. Then, annotations were extracted from \texttt{span} elements, classified into \texttt{PER}, \texttt{LOC} or \texttt{ORG} based on the \texttt{class} attribute. These elements were linked to potential Wikidata entities via the \texttt{owl:sameAs} property found in the \texttt{meta} elements of the HTML \texttt{head}. 

In the case of DZ, entity annotations were located within \texttt{link} elements, categorized into \texttt{PER}, \texttt{LOC} or \texttt{WORK} based on regular markup patterns. For both editions, if a mention was not linked to Wikidata it was tagged as NIL. Given the ongoing development of both digital editions, the structure of the annotations is subject to change, necessitating adaptable pre-processing strategies in future research.

Each SDE required a different sampling approach due to their distinct structures. From DZ, we selected two sections of the Zibaldone: pages 1000--2001 and pages 2700--4000, written in 1821 and 1823 respectively, corresponding to the years of highest productivity for the poet. For AMD, we sampled the first paragraph of each document, an approach chosen because AMD documents vary significantly in length (from short letters to lengthy political speeches), and first paragraphs typically introduce the main topics and key entities.

To ensure quality and balance, we excluded texts whose length deviated significantly from the standard distribution (beyond 2 standard deviations) and retained only documents containing at least one entity. This process produced 1,050 items from DZ and 1,061 from AMD. We then created train, validation, and test splits using a 70/15/15 ratio with stratified sampling based on creation year to maintain similar chronological distributions across all splits. The source code used for extracting this dataset and the documentation for using ENEIDE is available on Github.\footnote{\href{https://github.com/sntcristian/ENEIDE}{github.com/sntcristian/ENEIDE}}

\subsection{Quality Control}

Domain experts evaluated annotation quality using 100 randomly sampled documents from each dataset. The evaluation involved four expert scholars divided into two teams: one team with expertise in Italian literature (two PhD holders) and another in history (two PhD holders), coordinated by one of the authors to solve ambiguous cases. The evaluation followed these guidelines:

\begin{itemize}
\item Only annotate entities of type person, location, organization, and literary work that are referenced by proper names (i.e., exclude common nouns);
\item Do not annotate nested entities; always annotate the longest surface form (e.g., \textit{[Government of Italy]} rather than \textit{[Government]} of \emph{[Italy]});
\item Select the Wikidata identifier that best matches the entity in its historical and textual context;
\item Consider figures of speech such as metonymy (e.g., literary works referred to by their creator's name).
\end{itemize}

Table~\ref{tab:tab1} presents the results. DZ annotations achieved high quality with an F1 score of 95.6. AMD scored lower (79.8 F1), primarily due to missing annotations, as reflected in its recall of 70.6.

\begin{table}[h]
\centering
    \begin{tabular}{lll}
    \toprule
     \textbf{Statistic} & \textbf{DZ} & \textbf{AMD} \\
    \midrule
    Annotations in samples & 403 & 215 \\
    Wrong annotations & 13 & 18 \\
    Missing annotations & 10 & 64 \\
    Precision & 96.8 & 91.6 \\
    Recall & 94.4 & 70.6 \\
    F1 & 95.6 & 79.8 \\
    \bottomrule
    \end{tabular}
\caption{Quality assessment statistics for DZ and AMD samples.}
\label{tab:tab1}
\end{table}

To improve AMD's annotation coverage, we developed a semi-automatic enhancement pipeline comprising four steps: (i) extracting frequent entity mentions with Wikidata IDs and validating them with domain experts; (ii) automatically annotating previously missed mentions of validated entities throughout the text using exact string matching; (iii) applying the Italian StanzaNLP NER model\footnote{\href{https://stanfordnlp.github.io/stanza/ner_models.html}{stanfordnlp.github.io/stanza/ner\_models.html}} to identify remaining unannotated entities; and (iv) conducting final expert validation of automatically added annotations. This approach increased annotation completeness while maintaining annotation quality through expert oversight at each stage.

\begin{table*}[h]
\centering
\begin{tabular}{lcc}
\toprule
& \textbf{DZ} & \textbf{AMD} \\
\midrule
Documents (train) & 735 & 743 \\
Documents (dev) & 157 & 159 \\
Documents (test) & 158 & 160 \\
\midrule
Tokens per Doc (train) & 246.6 & 125.9 \\
Tokens per Doc (dev) & 253.2 & 119.3 \\
Tokens per Doc (test) & 248.7 & 119 \\
\midrule
Annotations (train) & 2,935 & 2,766 \\
Annotations (dev) & 727 & 604 \\
Annotations (test) & 617 & 657 \\
\midrule
\% Identifiers Overlap (train+dev vs test) & 93.19 & 75.38 \\
\bottomrule
\end{tabular}
\caption{Number of documents, average tokens, total annotations, and Wikidata entity overlap percentages in ENEIDE.}
\label{tab:tab2}
\end{table*}

\begin{table*}[h]
\centering
\begin{tabular}{lcccccccccc}
\toprule
& \multicolumn{5}{c}{\textbf{DZ}} & \multicolumn{5}{c}{\textbf{AMD}} \\
\cmidrule(lr){2-6} \cmidrule(lr){7-11}
\textbf{Split} & \textbf{PER} & \textbf{LOC} & \textbf{WORK} & \textbf{NIL} & \textbf{IDs} & \textbf{PER} & \textbf{LOC} & \textbf{ORG} & \textbf{NIL} & \textbf{IDs} \\
\midrule
train & 1,661 & 488 & 786 & 182 & 623 & 759 & 940 & 1,067 & 64 & 583 \\
dev & 375 & 149 & 203 & 74 & 276 & 158 & 226 & 206 & 13 & 203 \\
test & 318 & 130 & 169 & 42 & 241 & 194 & 190 & 205 & 9 & 238 \\
\bottomrule
\end{tabular}
\caption{Annotation counts by entity type (PER=person, LOC=location, WORK=literary work, ORG=organization), NIL entities, and unique Wikidata identifiers (IDs) in each split.}
\label{tab:tab3}
\end{table*}

\subsection{Dataset Statistics}

Tables~\ref{tab:tab2} and~\ref{tab:tab3} summarize key statistics for both corpora. Table~\ref{tab:tab2} shows document counts and number of annotations, while Table~\ref{tab:tab3} provides a fine-grained breakdown by entity type, reporting the number of NIL entities and unique Wikidata identifiers in each split. NIL entities constitute a marginal fraction of the total annotations: 7\% in DZ and 2\% in AMD. A notable difference between the two corpora is the entity overlap between training and test sets: 93.19\% of unique entities in the DZ train and dev splits also appear in the test set, whereas AMD exhibits lower overlap (75.38\%). This difference reflects the distinct natures of the two sources: DZ's philosophical reflections repeatedly reference a core set of classical authors and works, while AMD's political writings engage with a more diverse and time-specific set of contemporary figures and organizations. In terms of scale, ENEIDE is comparable to other medium-sized corpora for historical NERL, such as AJMC~\citep{ajmc_2024} and TopRes19th~\citep{ardanuy2022dataset}.

\section{Experiments}
\label{sec:experiments}

To establish baseline performance on ENEIDE and demonstrate its utility for evaluating NERL systems, we conducted experiments using SoTA models for both NER and EL tasks. All models were chosen for being open-source and versioned, thus allowing better reproducibility of our results.

\subsection{Named Entity Recognition}
\label{sec:NER_results}

For NER, we evaluated four architectures. First, we considered three instruction-tuned Large Language Models (LLMs) using zero-shot prompting: LLaMA 3.1-8B\footnote{\href{https://huggingface.co/meta-llama/Llama-3.1-8B-Instruct}{meta-llama/Llama-3.1-8B-Instruct}}, a general-purpose multilingual instruction-tuned model; Minerva-7B\footnote{\href{https://huggingface.co/sapienzanlp/Minerva-7B-instruct-v1.0}{sapienzanlp/Minerva-7B-instruct-v1.0}}, an Italian-focused instruction-tuned model; and Ministral-8B\footnote{\href{https://huggingface.co/mistralai/Ministral-8B-Instruct-2410}{mistralai/Ministral-8B-Instruct-2410}}, a multilingual instruction-tuned model from the Mistral family. Each model was prompted to identify and classify named entities according to the entity types present in each dataset (person, location, organization, and literary work) and to return them in a list formatted as JSON. A detailed overview of the prompts used is given in Appendix~\ref{sec:ner_prompts}.

In addition, to compare LLMs used in a zero-shot setting with a smaller model trained on ENEIDE, we decided to fine-tune a GLiNER model~\cite{zaratiana2023gliner}  already pre-trained for universal NER on Italian\footnote{\href{https://huggingface.co/DeepMount00/universal_ner_ita}{huggingface.co/DeepMount00/universal\_ner\_ita}} using the full ENEIDE training set (DZ and AMD).

Table~\ref{tab:ner_results} presents the results. GliNER (fine-tuned) achieved the best performance across both datasets, with F1 scores of 0.782 on DZ and 0.876 on AMD. The slightly lower performance on DZ reflects the challenges posed by references to \texttt{WORK} entities, where the model achieved an F1 score of 0.592 using a \textit{strict match} criterion and 0.724 using \textit{relaxed match} (at least one token overlap). 

All the LLMs, despite having a larger number of parameters, exhibited very low precision and recall, suggesting difficulties in correctly interpreting instructions for NER when prompted with domain-specific and historical data. Overall, the moderate F1 scores across all zero-shot models highlight the difficulty of historical NER and the need for specialized approaches.

\begin{table}[h]
\centering
\resizebox{\linewidth}{!}{
\begin{tabular}{llccc}
\toprule
\textbf{Dataset} & \textbf{Model} & \textbf{Precision} & \textbf{Recall} & \textbf{F1} \\
\midrule
\multirow{3}{*}{DZ} & LLaMA 3.1-8B & 0.438 & 0.376 & 0.405 \\
 & Minerva-7B & 0.531 & 0.028 & 0.052 \\
 & Ministral-8B & 0.351 & 0.329 & 0.340 \\
& GLiNER (fine-tuned) & \bf 0.851 & \bf 0.723 & \bf 0.782 \\
\midrule
\multirow{3}{*}{AMD} & LLaMA 3.1-8B & 0.719 & 0.603 & 0.656 \\
 & Minerva-7B & 0.659 & 0.138 & 0.228 \\
 & Ministral-8B & 0.678 & 0.530 & 0.595 \\
 & GLiNER (fine-tuned) & \bf 0.887 & \bf 0.866 & \bf 0.876 \\
\bottomrule
\end{tabular}
}
\caption{NER results on DZ and AMD test sets using zero-shot instruction-tuned LLMs.}
\label{tab:ner_results}
\end{table}

\subsection{Entity Linking}

For EL, we evaluated three publicly available general-purpose models trained on Wikipedia: BLINK-ita\footnote{\href{https://github.com/rpo19/pozzi_aixia_2023}{github.com/rpo19/pozzi\_aixia\_2023}}~\citep{pozzi2023named}, an Italian adaptation of the BLINK model~\citep{wu2020scalable}; mGENRE\footnote{\href{https://github.com/facebookresearch/GENRE}{github.com/facebookresearch/GENRE}}~\citep{de2022multilingual}, a multilingual generative entity retrieval model; and BELA\footnote{\href{https://github.com/facebookresearch/BELA}{github.com/facebookresearch/BELA}}~\citep{plekhanov2023multilingual}, a multilingual bi-encoder for EL. We evaluated all models in the entity disambiguation task, i.e. we used ground-truth entity spans from the test sets and evaluated the models' ability to correctly link them to Wikidata identifiers. 

Table~\ref{tab:el_results} presents the results. BELA achieved the best performance on DZ (0.598 accuracy), while mGENRE performed best on AMD (0.689 accuracy). The higher accuracy scores on AMD compared to DZ can be attributed to two factors: first, AMD entities are predominantly contemporary political figures and organizations that are well-represented in Wikipedia training data; second, DZ contains numerous classical and literary references that may be underrepresented or absent in standard Wikipedia-based training corpora. These results demonstrate that while general-purpose EL models achieve reasonable performance on recent texts containing factual information, they struggle with specialized literary and historical content as in DZ.

\begin{table}[h]
\centering
\begin{tabular}{llc}
\toprule
\textbf{Dataset} & \textbf{Model} & \textbf{Accuracy} \\
\midrule
\multirow{3}{*}{DZ} & mGENRE & 0.579 \\
 & BLINK-ita & 0.502 \\
 & BELA & \textbf{0.598} \\
\midrule
\multirow{3}{*}{AMD} & mGENRE & \textbf{0.689} \\
 & BLINK-ita & 0.643 \\
 & BELA & 0.676 \\
\bottomrule
\end{tabular}
\caption{Entity disambiguation results on DZ and AMD test sets using EL systems trained on Wikipedia.}
\label{tab:el_results}
\end{table}

\section{Discussion}
\label{sec:discussion}

\begin{figure*}[h]
    \centering
    \includegraphics[width=1\textwidth]{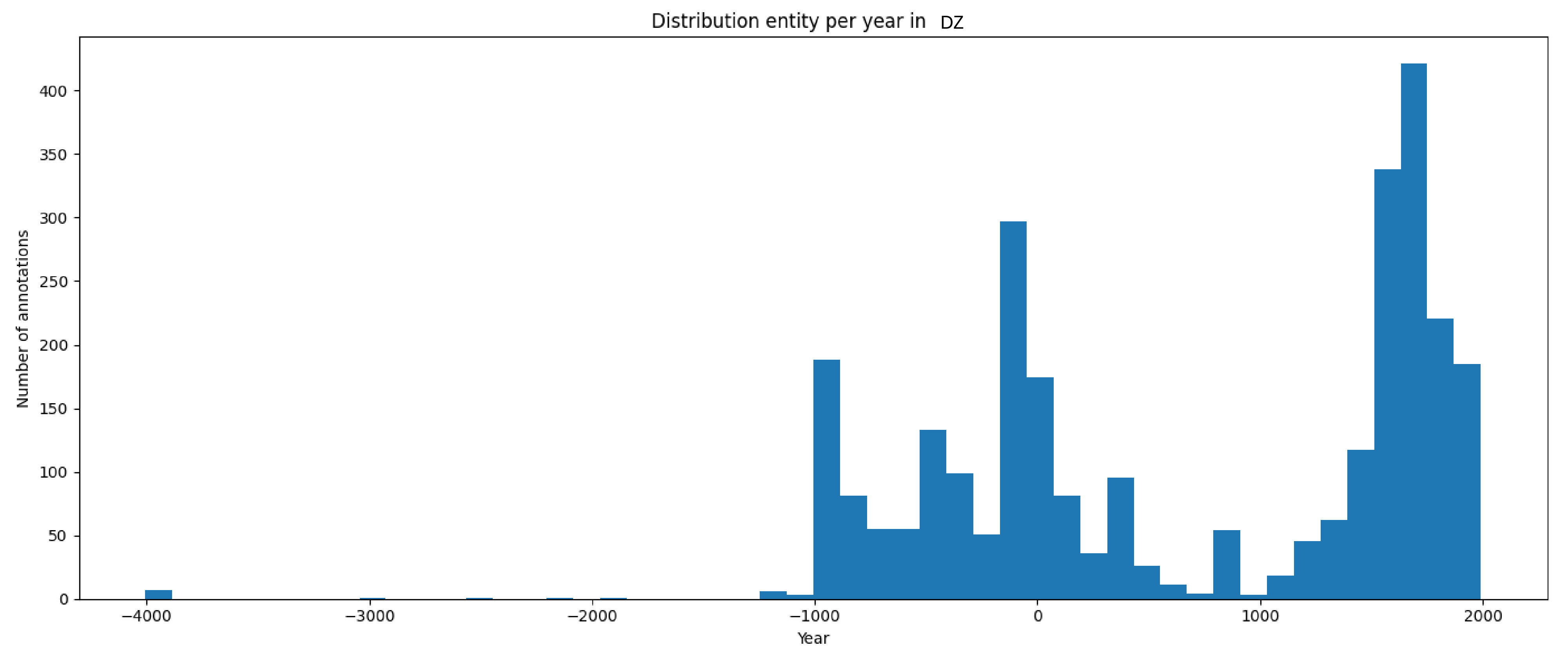}
    \caption{Temporal distribution of named entities in the DZ corpus, showing references spanning from classical antiquity to the early modern period.}
    \label{fig:dz_temporal}
\end{figure*}

\begin{figure*}[h]
    \centering
    \includegraphics[width=1\textwidth]{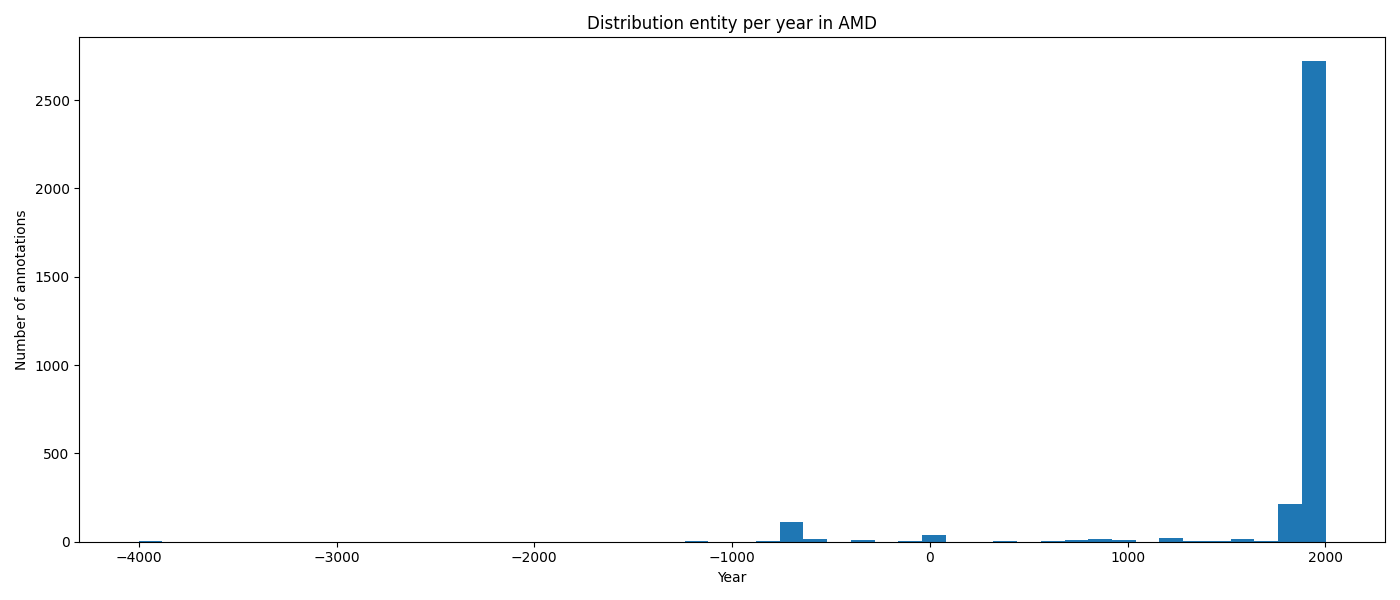}
    \caption{Temporal distribution of named entities in the AMD corpus, showing references concentrated in the 20th-century Italian history.}
    \label{fig:amd_temporal}
\end{figure*}

ENEIDE aims to be a valuable resource for NLP researchers in the Italian DH community. Since every text is annotated with temporal information related to its creation date, this dataset can be used effectively for training and evaluating algorithms for diachronic EL~\citep{agarwal2018dianed}. Figures~\ref{fig:dz_temporal} and~\ref{fig:amd_temporal} show the temporal distribution of entities in both dataset partitions, highlighting their diachronic coverage. This distribution was obtained by querying Wikidata for the earliest recorded date of each entity in the corpus and plotting a frequency distribution of annotations by year.

The DZ partition (Figure~\ref{fig:dz_temporal}) is particularly valuable for scholars focused on recognizing and linking literary entities and cultural references. By providing numerous annotations of classical works and literary authors across many historical periods (spanning from ancient Greece and Rome through the Renaissance to the early modern period) DZ serves as a challenging benchmark for NLP systems tailored to the humanistic domain. The temporal distribution reveals a concentration of entities from classical antiquity and the early modern period, reflecting Leopardi's philosophical interests and the foundational texts that influenced his thinking.

The AMD partition (Figure~\ref{fig:amd_temporal}) enables improvement and verification of systems designed for detecting and disambiguating entities in the political domain. This partition contains rich annotations of historical figures and political organizations that shaped the 20th-century Italian history, with entity distributions concentrated in the mid-20th century, corresponding to Moro's active political career. The resource can therefore support historical studies that leverage EL systems to analyze the evolution of social networks and their relationship with political and historical dynamics~\citep{nabiafjadi2021social}.

Beyond these domain-specific applications, ENEIDE's multi-domain nature makes it suitable for evaluating cross-domain generalization capabilities of NERL systems. The baseline experiments presented in Section~\ref{sec:experiments} demonstrate that current SoTA models struggle with historical Italian texts, particularly in the literary domain, suggesting substantial room for improvement through specialized training or domain adaptation techniques. The dataset's inclusion of NIL entities further enables research on entity discovery and KG completion~\citep{zhu2023learn}, as these non-linkable entities represent gaps in existing knowledge resources that could be addressed through automated methods.

\section{Conclusion}
\label{sec:conclusion}

This paper presented ENEIDE, the first multi-domain, publicly available dataset for NERL in historical Italian with training, development, and test splits. Comprising 2,111 documents with over 8,000 entity annotations spanning two centuries and two distinct domains, ENEIDE addresses a critical gap in language resources for historical Italian NLP. 

We presented a methodology for semi-automatic dataset extraction from SDEs, including quality control procedures and annotation enhancement techniques that can be potentially reproduced to other digital editions, given a certain degree of structural similarity with the ones that we treated in this work. Our baseline experiments demonstrate that current SoTA NERL systems struggle with historical Italian texts, particularly in the literary domain, with F1 scores ranging from 0.340 to 0.876 for NER and accuracy scores from 0.502 to 0.689 for EL. These results highlight both the challenge posed by historical texts and the value of ENEIDE as a benchmark for developing specialized NERL systems.

The dataset's diachronic coverage, spanning from classical antiquity to 20th-century Italy, makes it particularly suitable for research on temporal entity disambiguation and cross-domain generalization. The inclusion of domain-specific entity types (literary works and organizations) alongside general types (persons and locations) further broadens its applicability to specialized NLP tasks in DH.

We release ENEIDE under a CC BY-NC-SA 4.0 license to encourage its adoption by the research community and hope it will stimulate further work on historical Italian NLP and the reuse of annotated resources in the DH as training and evaluation data for language technologies.

\section{Limitations}
\label{sec:limitations}

Despite its contributions, ENEIDE has several limitations that should be considered when using this resource. First, due to its limited representativeness of historical Italian varieties, automatic systems trained exclusively on this resource may learn domain-specific biases if not carefully designed and tested for generalizability. The dataset covers only two authors and two specific domains (philosophical-literary and political-legal), which may not reflect the full linguistic and stylistic diversity of historical Italian texts.

Second, the annotation enhancement pipeline applied to AMD, while improving coverage, may have introduced some inconsistencies compared to the original expert-curated DZ annotations. Although the enhancement process included expert validation steps, the semi-automatic nature of the additions may result in subtle quality differences between the two partitions.

Finally, the dataset currently focuses on Italian texts with limited multilingual coverage (some excerpts in French, Latin, and Greek appear in DZ). This limits its applicability for multilingual or cross-lingual NERL research.

Nonetheless, ENEIDE is an actively maintained resource, and future work may address these limitations by extending the dataset with additional authors, genres, and time periods to broaden its scope and enhance its reusability. We also plan to explore the integration of additional SDEs to increase domain diversity and temporal coverage.

\section*{Acknowledgments}

We acknowledge the contribution of the development team of Aldo Moro Digitale~\citep{barzaghi_amd_2025} and DigitalZibaldone~\citep{stoyanova_remediating_2014} for providing the original HTML files from which ENEIDE was extracted. This publication is based upon work from COST Action CA24121 Knowledge Graphs in the Era of Large Language Models (KGELL), supported by COST (European Cooperation in Science and Technology, \href{https://www.cost.eu}{www.cost.eu})."

\section*{Declaration on Generative AI}

During the preparation of this work, the authors used Claude (\href{https://www.anthropic.com/claude/sonnet}{Sonnet 4.5}) to perform a proofreading of the initial draft. After using this tool, the authors reviewed and edited the content as needed to take full responsibility for the publication's content. No AI tool was used for designing, interpreting and analysing the research itself.

\section*{References}

\bibliographystyle{lrec2026-natbib}
\bibliography{lrec2026-bibliography}

\begin{appendix}
\label{sec:appendix}

\section{Prompts for NER with Instruction-Tuned LLM}
\label{sec:ner_prompts}
This appendix details the prompts used to perform NER with instruction-tuned LLMs, as presented in Section~\ref{sec:NER_results}. Each LLM was prompted to identify and classify named entities according to the entity types present in each dataset (person, location, organization, and literary work) and to return them in a list formatted as JSON. 

\subsection{Prompt for DZ}
\begin{quote}
\textbf{System Prompt:} You are a philologist with expert knowledge of Italian literature. Your task is to annotate references to Persons, Locations, and Literary Works within historical texts. You generate structured responses in JSON format. Do not write Python code.

\textbf{User Prompt:} Extract the references to named entities of type ``person'', ``location'', and ``work'' within the input text, taken from the collection ``Zibaldone di pensieri'' by the poet and philologist Giacomo Leopardi (1817--1832). Extract the entities in the response using a JSON format as in the provided example.

Example Input: ``La Divina Commedia venne scritta da Dante a Firenze''.

Example Output:
\begin{verbatim}
[["Divina Commedia", "work"],
["Dante", "person"],
["Firenze", "location"]]
\end{verbatim}

Input Text: \texttt{\{Document\}}
\end{quote}

\subsection{Prompt for AMD}
\begin{quote}
\textbf{System Prompt:} You are a political scientist with expert knowledge of Italian politics. Your task is to annotate references to Persons, Locations, and Organizations within historical texts. You generate structured responses in JSON format. Do not write Python code.

\textbf{User Prompt:} Extract the references to named entities of type ``person'', ``location'', and ``organization'' within the input text, taken from the writings of the Italian politician Aldo Moro (1916--1978). Extract the entities in the response using a JSON format as in the provided example.

Example Input: ``Nel luglio Togliatti si trasfer\`{i} a Mosca dove partecip\`{o} al VI congresso dell'Internazionale Comunista.''

Example Output:
\begin{verbatim}
[["Togliatti", "person"],
["Mosca", "location"],
["Internazionale Comunista", "organization"]]
\end{verbatim}

Input Text: \texttt{\{Document\}}
\end{quote}

\end{appendix}
\end{document}